\documentclass[]{article}
\usepackage[letterpaper]{geometry}
\usepackage{mtsummit2017}
\usepackage{times}
\usepackage{url}
\usepackage{latexsym}
\usepackage{natbib}
\usepackage{layout}
\usepackage{graphicx} 
\usepackage{enumitem}
\usepackage{multirow}
\graphicspath{ {images/} }
\usepackage{colortbl}
\usepackage[T1]{fontenc}
\begin{document}

\title{Enabling Multi-Source Neural Machine Translation By Concatenating Source Sentences In Multiple Languages}  

\author{\name{\bf Raj Dabre} \hfill  \addr{raj@nlp.ist.i.kyoto-u.ac.jp}\\
        \addr{Graduate School of Informatics, Kyoto University, Kyoto, Japan}
\AND
        \name{\bf Fabien Cromieres} \hfill \addr{fabien@pa.jst.jp}\\ 
        \addr{Japan Science and Technology Agency, Saitama, Japan}
\AND
       \name{\bf Sadao Kurohashi} \hfill \addr{kuro@i.kyoto-u.ac.jp}\\
        \addr{Graduate School of Informatics, Kyoto University, Kyoto, Japan}
}

\maketitle
\pagestyle{empty}

\begin{abstract}
  In this paper, we explore a simple solution to ``Multi-Source Neural Machine Translation" (MSNMT) which only relies on preprocessing a N-way multilingual corpus without modifying the Neural Machine Translation (NMT) architecture or training procedure. We simply concatenate the source sentences to form a single long multi-source input sentence while keeping the target side sentence as it is and train an NMT system using this preprocessed corpus. We evaluate our method in resource poor as well as resource rich settings and show its effectiveness (up to 4 BLEU using 2 source languages and up to 6 BLEU using 5 source languages). We also compare against existing methods for MSNMT and show that our solution gives competitive results despite its simplicity. We also provide some insights on how the NMT system leverages multilingual information in such a scenario by visualizing attention.
\end{abstract}

\section{Introduction}
Multi-Source Machine Translation (MSMT) \cite{och2001statistical} is an approach that allows one to leverage source sentences in multiple languages to improve the translations to a target language. Typically N-way (or N-lingual) corpora are used for MSMT. N-way corpora are those in which translations of the same sentence exist in N different languages. This setting is realistic and has many applications. For example, the European Parliament maintains its proceedings in 21 languages. In Spain, international news companies write news articles in English as well as Spanish. One can now utilize the same sentence written in two different languages like Spanish and English to translate to a third language like Italian by utilizing a large English-Spanish-Italian trilingual corpus.\\
Neural machine translation (NMT) \cite{DBLP:journals/corr/BahdanauCB14,DBLP:journals/corr/ChoMGBSB14,DBLP:journals/corr/SutskeverVL14} enables one to train an end-to-end system without the need to deal with word alignments, translation rules and complicated decoding algorithms, which are a characteristic of phrase based statistical machine translation (PBSMT) systems. However, it is reported that NMT works better than PBSMT only when there is an abundance of parallel corpora. In a low resource scenario, vanilla NMT is either worse than or comparable to PBSMT \cite{DBLP:conf/emnlp/ZophYMK16}.\\
Multi-source Neural Machine translation (MSNMT) involves using NMT for MSMT. Two major approaches for Multi-Source NMT (MSNMT) have been explored, namely the multi-encoder (ME/me) \cite{zoph-knight:2016:N16-1} and multi-source ensembling (ENS/ens) \cite{garmash-monz:2016:COLING,DBLP:conf/emnlp/FiratSAYC16}. The multi-encoder approach involves extending the vanilla NMT architecture to have an encoder for each source language leading to larger models. On the other hand, the ensembling approach is simpler since it involves training multiple bilingual NMT models each with a different source language but the same target language.\\
We have discovered that there is an even simpler way to do MSNMT. We explore a new simplified end-to-end method that avoids the need to modify the NMT architecture as well as the need to learn an ensemble function. We simply concatenate the source sentences leading to a parallel corpus where the source side is a long multilingual sentence and the target side is a single sentence which is the translation of the aforementioned multilingual sentence. This corpus is then fed to any NMT training pipeline whose output is a multi-source NMT model.\\
The main contributions of this paper are as follows:
\begin{itemize}[topsep=0pt,itemsep=-1ex,partopsep=1ex,parsep=1ex]
    \item Exploring a simple preprocessing step that allows for Multi-Source NMT (MSNMT) without any change to the NMT architecture\footnote{One additional benefit of our approach is that any NMT architecture can be used, be it attention based or hierarchical NMT.}.
    \item An exhaustive study of how the approach works in a resource poor as well as a resource rich setting.
    \item An analysis of how gains in the translation quality are correlated with language similarity in a multi-source scenario.
    \item An empirical comparison of our approach against two existing methods \cite{zoph-knight:2016:N16-1,DBLP:conf/emnlp/FiratSAYC16} for MSNMT.
    \item An analysis of how NMT gives more importance to certain linguistically closer languages while doing multi-source translation by visualizing attention vectors.
\end{itemize}


\section{Related Work}
One of the first studies on multi-source MT \cite{och2001statistical} explored how word based SMT systems would benefit from multiple source languages. Although effective, it suffered from a number of limitations that classic word and phrase based SMT systems do including the inability to perform end-to-end training. The work on multi-encoder multi source NMT \cite{zoph-knight:2016:N16-1} is the first multi-source NMT approach which focused on utilizing French and German as source languages to translate to English. However their method led to models with substantially larger parameter spaces and they did not experiment with many languages. Moreover, since the encoders for each source language are separate it is difficult to explore how the source languages contribute towards the improvement in translation quality. Multi-source ensembling using a multilingual multi-way NMT model \cite{DBLP:conf/emnlp/FiratSAYC16} is an end-to-end approach but requires training a very large and complex NMT model. The work on multi-source ensembling which uses separately trained single source models \cite{garmash-monz:2016:COLING} is comparatively simpler in the sense that one does not need to train additional NMT models but the approach is not truly end-to-end since it needs an ensemble function to be learned. This method also helps eliminates the need for N-way corpora which allows one to exploit bilingual corpora which are larger in size. In all cases one ends up with either one large model or many small models for which an ensemble function needs to be learned.\\
Other related works include Transfer Learning \cite{DBLP:conf/emnlp/ZophYMK16} and Zero Shot NMT \cite{gnmt16multi} which help improve NMT performance for low resource languages. Finally it is important to note works that involve the creation of N-way corpora: United Nations (\cite{ZIEMSKI16.1195}), Europarl (\cite{koehn2005epc}), Ted Talks (\cite{cettoloEtAl:EAMT2012}), ILCI (\cite{JHA10.874}) and Bible (\cite{Christodouloupoulos2015}) corpora.

\section{Our Method}

\begin{figure}[t]
    \centering
  \centerline{\includegraphics[width=7.5cm,height=6cm]{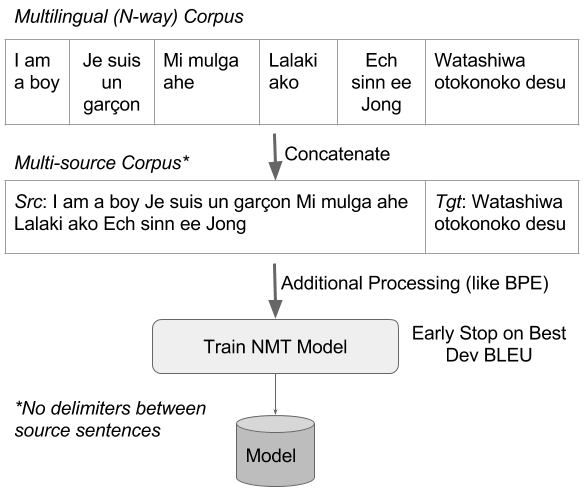}}
    \caption{The Multi-Source NMT Approach We Explored.}
    \label{msnmt}
\end{figure}

Refer to Figure~\ref{msnmt} for an overview of our method which is as follows: For each target sentence \textbf{concatenate the corresponding source sentences} leading to a parallel corpus where the source sentence is a very long sentence that conveys the same meaning in multiple languages. An example line in such a corpus would be: source: ``Hello Bonjour Namaskar Kamusta Hallo" and target:``konnichiwa". The 5 source languages here are English, French, Marathi, Filipino and Luxembourgish whereas the target language is Japanese. In this example each source sentence is a word conveying ``Hello" in different languages. Note that there are \textbf{no delimiters between the individual source sentences} since we expect the NMT system will figure out the sentence boundaries by itself. We romanize the Marathi and Japanese words for readability. Optionally, one can perform additional processing, like Byte Pair Encoding (BPE), to overcome data sparsity and eliminate the unknown word rate. Use the training corpus to learn an NMT model using any off the shelf NMT toolkit. The order of the sentences belonging to different languages is kept the same in the training, development and test sets.

\subsection{Other methods for comparison}
\subsubsection{Multi-Encoder Multi-Source Method}
This method was proposed by \cite{zoph-knight:2016:N16-1}. The main idea is to have an encoder for each source language and concatenate encoding information before feeding it to the decoder. We use the technique where attentions are computed for both source languages and feed this multi-source attention to the decoder to predict a target word.

\subsubsection{Multi-Source Ensembling Method}
This method was proposed by \cite{DBLP:conf/emnlp/FiratSAYC16} and it relies on a single multilingual NMT model with separate encoders and decoders for each source and target language. All encoders and decoders share a single attention mechanism. To perform multi-source translation the model is fed source sentences in different languages and the logits are averaged (ensembling) before computing softmax to predict a target word. Since training a multilingual-multiway model is difficult and time consuming to train we rely on separately trained models for each source language and ensemble them without learning an ensemble function.

\section{Experimental Settings}
\label{sec:settings}
All of our experiments were performed using an encoder-decoder NMT system with attention for the various baselines and multi-source experiments. In order to enable infinite vocabulary and reduce data sparsity we use the Byte Pair Encoding (BPE) based word segmentation approach \cite{DBLP:journals/corr/SennrichHB15}. However we perform a slight modification to the original code where instead of specifying the number of merge operations manually we specify a desired vocabulary size and the BPE learning process automatically stops after it learns enough rules to obtain the prespecified vocabulary size. We prefer this approach since it allows us to learn a minimal model and it resembles the way Google's NMT system \cite{DBLP:journals/corr/WuSCLNMKCGMKSJL16} works with the Word Piece Model (WPM) \cite{DBLP:conf/icassp/SchusterN12}. We evaluate our models using the standard BLEU \cite{Papineni:2002:BMA:1073083.1073135} metric\footnote{This is computed by the multi-bleu.pl script, which can be downloaded from the public implementation of Moses \cite{conf/acl/KoehnHBCFBCSMZDBCH07}.} on the translations of the test set. Baseline models are single source models.

\subsection{Languages and Corpora Settings}
\begin{table*}[h]
\small
\begin{center}
\begin{tabular}{|l|l|l|l|l|}
\hline
corpus type & Languages          & train  & dev2010 & tst2010/tst2013 \\ \hline
3 lingual   & Fr, De, En         & 191381 & 880     & 1060/886        \\ \hline
4 lingual   & Fr, De, Ar, En     & 84301  & 880     & 1059/708        \\ \hline
5 lingual   & Fr, De, Ar, Cs, En & 45684  & 461     & 1016/643        \\ \hline
\end{tabular}
\end{center}
\caption{Statistics for the the N-lingual corpora extracted from the IWSLT corpus for the languages French (Fr), German (De), Arabic (Ar), Czech (Cs) and English (En)}
\label{iwsltstats}
\end{table*}

All of our experiments were performed using the publicly available ILCI\footnote{This was used for the Indian Languages MT task in ICON 2014\footnote{http://ltrc.iiit.ac.in/icon/2014} and 2015\footnote{http://ltrc.iiit.ac.in/icon2015/}.} (\cite{JHA10.874}), United Nations\footnote{https://conferences.unite.un.org/uncorpus} (\cite{ZIEMSKI16.1195}) and IWSLT\footnote{https://wit3.fbk.eu/mt.php?release=2016-01} (\cite{cettolo2015iwslt}) corpora.\\
The ILCI corpus is a 6-way multilingual corpus spanning the languages Hindi, English, Tamil, Telugu, Marathi and Bengali was provided as a part of the task. The target language is Hindi and thus there are 5 source languages. The training, development and test sets contain 45600, 1000 and 2400 6-lingual sentences respectively\footnote{In the task there are 3 domains: health, tourism and general. However, we focus on the general domain in which half the corpus comes from the health domain the other half comes from the tourism domain.}. Hindi, Bengali and Marathi are Indo-Aryan languages, Telugu and Tamil are Dravidian languages and English is a European language. In this group English is the farthest from Hindi, grammatically speaking, whereas Marathi is the closest to it. Morphologically speaking, Bengali is closer to Hindi compared to Marathi (which has agglutinative suffixes) but Marathi and Hindi share the same script and they also share more cognates compared to the other languages. It is natural to expect that translating from Bengali and Marathi to Hindi should give Hindi sentences of higher quality as compared to those obtained by translating from the other languages and thus using these two languages as source languages in multi-source approaches should lead to significant improvements in translation quality. We verify this hypothesis by exhaustively trying all language combinations.\\
The IWSLT corpus is a collection of 4 bilingual corpora spanning 5 languages where the target language is English: French-English (234992 lines), German-English (209772 lines), Czech-English (122382 lines) and Arabic-English (239818 lines). Linguistically speaking French and German are the closest to English followed by Czech and Arabic. In order to obtain N-lingual sentences we only keep the sentence pairs from each corpus such that the English sentence is present in all the corpora. From the given training data we extract trilingual (French, German and English), 4-lingual (French, German, Arabic and English) and 5-lingual corpora. Similarly we extract 3, 4 and 5 lingual development and test sets. The IWSLT corpus (downloaded from the link given above) comes with a development set called dev2010 and test sets named tst2010 to tst2013 (one for each year from 2010 to 2013). Unfortunately only the tst2010 and tst2013 test sets are N-lingual. Refer to Table~\ref{iwsltstats} which contains the number of lines of training, development and test sentences we extracted.\\
The UN corpus spans 6 languages: French, Spanish, Arabic, Chinese, Russian and English. Although there are 11 million 6-lingual sentences we use only 2 million for training since our purpose was not to train the best system but to show that our method works in a resource rich situation as well. The development and test sets provided contain 4000 lines each and are also available as 6-lingual sentences. We chose English to be the target language and focused on Spanish, French, Arabic and Russian as source languages. Due to lack of computational facilities we only worked with the following source language combinations: French and Spanish, French and Russian, French and Arabic and Russian and Arabic.

\subsection{NMT Systems and Model Settings}
For training various NMT systems, we used the open source KyotoNMT toolkit\footnote{https://github.com/fabiencro/knmt} \cite{cromieres-EtAl:2016:WAT2016}. KyotoNMT implements an Attention based Encoder-Decoder \cite{DBLP:journals/corr/BahdanauCB14} with slight modifications to the training procedure. We modify the NMT implementation in KyotoNMT to enable multi encoder multi source NMT \cite{zoph-knight:2016:N16-1}. Since the NMT model architecture used in \cite{zoph-knight:2016:N16-1} is slightly different from the one in KyotoNMT, the multi encoder implementation is not identical (but is equivalent) to the one in the original work. For the rest of the paper ``baseline" systems indicate single source NMT models trained on bilingual corpora. We train and evaluate the following NMT models:
\begin{itemize}[topsep=0pt,itemsep=-1ex,partopsep=1ex,parsep=1ex]
    \item One source to one target.
    \item N source to one target using our proposed multi source approach.
    \item N source to one target using the multi encoder multi source approach \cite{zoph-knight:2016:N16-1}.
    \item N source to one target using the multi source ensembling approach that late averages\footnote{Late averaging implies averaging the logits of multiple decoders before computing softmax to predict the target word.} \cite{DBLP:conf/emnlp/FiratSAYC16} N one source to one target models\footnote{In the original work a single multilingual multiway NMT model was trained and ensembled but we train separate NMT models for each source language.}.
\end{itemize}

The model and training details are as follows:
\begin{itemize}[topsep=0pt,itemsep=-1ex,partopsep=1ex,parsep=1ex]
    \item BPE vocabulary size: 8k\footnote{We also try vocabularies of size 16k and 32k but they take longer to train and overfit badly in a low resource setting} (separate models for source and target) for ILCI and IWSLT corpora settings and 16k for the UN corpus setting. When training the BPE model for the source languages we learn a single shared BPE model. In case of languages that use the same script it allows for cognate sharing thereby reducing the overall vocabulary size.
    \item Embeddings: 620 nodes
    \item RNN (Recurrent Neural Network) for encoders and decoders: LSTM with 1 layer, 1000 nodes output. Each encoder is a bidirectional RNN.
    \item In the case of multiple encoders, one for each language, each encoder has its own separate vocabulary.
    \item Attention: 500 nodes hidden layer. In case of the multi encoder approach there is a separate attention mechanism per encoder.
    \item Batch size: 64 for single source, 16 for 2 sources and 8 for 3 sources and above for IWSLT and ILCI corpora settings. 32 for single source and 16 for 2 sources for the UN corpus setting.
    \item Training steps: 10k\footnote{We observed that the models start overfitting around 7k-8k iterations} for 1 source, 15k for 2 source and 40k for 5 source settings when using the IWSLT and ILCI corpora. 200k for 1 source and 400k for 2 source for the UN corpus setting to ensure that in both cases the models get saturated with respect to heir learning capacity.
    \item Optimization algorithms: Adam with an initial learning rate of 0.01
    \item Choosing the best model: Evaluate the model on the development set and select the one with the best BLEU \cite{Papineni:2002:BMA:1073083.1073135} after reversing the BPE segmentation on the output of the NMT model.
    \item Beam size for decoding: 16\footnote{We performed evaluation using beam sizes 4, 8, 12 and 16 but found that the differences in BLEU between beam sizes 12 and 16 are small and gains in BLEU for beam sizes beyond 16 are insignificant}
\end{itemize}

We train and evaluate the following NMT models using the ILCI corpus:
\begin{itemize}[topsep=0pt,itemsep=-1ex,partopsep=1ex,parsep=1ex]
    \item One source to one target: 5 models (Baselines)
    \item Two source to one target: 10 models (5 source languages, choose 2 at a time)
    \item Five source to one target: 1 model
\end{itemize}

In this setting, we also calculate the \textbf{\textit{lexical similarity}}\footnote{https://en.wikipedia.org/wiki/Lexical\_similarity} between the languages involved in using the Indic NLP Library\footnote{http://anoopkunchukuttan.github.io/indic\_nlp\_library}. The objective behind this is to determine whether or not lexical similarity, which is also one of the indicators of linguistic similarity and hence translation quality \cite{DBLP:conf/emnlp/KunchukuttanB16}, is also an indicator of how well two source languages work together.\\

In the IWSLT corpus setting we did not try various combinations of source languages as we did in the ILCI corpus setting. We train and evaluate the following NMT models for each N-lingual corpus:
\begin{itemize}[topsep=0pt,itemsep=-1ex,partopsep=1ex,parsep=1ex]
    \item One source to one target: N-1 models (Baselines; 2 for the trilingual corpus, 3 for the 4-lingual corpus and 4 for the 5-lingual corpus)
    \item N-1 source to one target: 3 models (1 for trilingual, 1 for 4-lingual and 1 for 5-lingual)
\end{itemize}

Similarly for the UN corpus setting we only tried the following one source one target models: French-English, Russian-English, Spanish-English and Arabic-English. The two source combinations we tried were: French+Spanish, French+Arabic, French+Russian, Russian+Arabic. The target language is English.

\begin{table*}[t]
\small
\begin{center}
\setlength{\tabcolsep}{.09em}
\begin{tabular}{|c|c|c|c|c|c|c|c|c|c|c|c|c|c|c|c|c|}
\hline
\multirow{3}{*}{\begin{tabular}[c]{@{}c@{}}\textbf{Source}\\ \textbf{Language} \textbf{1}\end{tabular}} & \multicolumn{16}{c|}{\textbf{Source Language 2} {[}XX-Hi BLEU{]} \textit{XX-Hi \textit{sim}}}                                                                                                              \\ \cline{2-17}                                                                             & \multicolumn{4}{c|}{\textbf{En} {[}11.08{]} \textit{\tiny 0.20}} & \multicolumn{4}{c|}{\textbf{Mr} {[}24.60{]} \textit{\tiny 0.51}} & \multicolumn{4}{c|}{\textbf{Ta} {[}10.37{]} \textit{\tiny 0.30}} & \multicolumn{4}{c|}{\textbf{Te} {[}16.55{]} \textit{\tiny 0.42}} \\ \cline{2-17} 
                                                                             & our      & ens      & me       & \textit{sim}       & our      & ens      & me       & \textit{sim}       & our      & ens      & me       & \textit{sim}       & our      & ens      & me       & \textit{sim}       \\ \hline
\textbf{Bn} {[}19.14{]} \textit{\tiny 0.52}                                                        & \textbf{20.70}    & 19.45    & 19.10    & \textit{\tiny 0.18}    & 29.02    & \textbf{30.10}    & 27.33    & \textit{\tiny 0.46}    & 19.85    & \textbf{20.79}    & 18.26    & \textit{\tiny 0.30}    & 22.73    & \textbf{24.83}    & 22.14    & \textit{\tiny 0.39}    \\ \hline
\textbf{En} {[}11.08{]} \textit{\tiny 0.20}                                                        & \multicolumn{4}{c|}{-}                     & 25.56    & 23.06    & \textbf{26.01}    & \textit{\tiny 0.20}      & 14.03    & \textbf{15.05}    & 13.30    & \textit{\tiny 0.18}    & 18.91    & \textbf{19.68}    & 17.53    & \textit{\tiny 0.20}    \\ \hline
\textbf{Mr} {[}24.60{]} \textit{\tiny 0.51}                                                        & \multicolumn{4}{c|}{-}                     & \multicolumn{4}{c|}{-}                     & \textbf{25.64}    & 24.70    & 23.79    & \textit{\tiny 0.33}    & 27.62    & \textbf{28.00}    & 26.63    & \textit{\tiny 0.43}    \\ \hline
\textbf{Ta} {[}10.37{]} \textit{\tiny 0.30}                                                        & \multicolumn{4}{c|}{-}                     & \multicolumn{4}{c|}{-}                     & \multicolumn{4}{c|}{-}                     & 18.14    & \textbf{19.11}    & 17.34    & \textit{\tiny 0.38}    \\ \hline
All                                                                          & \multicolumn{5}{c|}{our: \textbf{31.56}}                       & \multicolumn{6}{c|}{ens: 30.29}                                  & \multicolumn{5}{c|}{me: 28.31}                         \\ \hline
\end{tabular}
\end{center}
\caption{\textbf{ILCI corpus results}: BLEU scores \label{ms1t} for two source to one target setting for all language combinations and for five source to one target using the ILCI corpus. The languages are Bengali (Bn), English (En), Marathi (Mr), Tamil (Ta), Telugu (Te) and Hindi (Hi). Each language is accompanied by the BLEU score for translating to Hindi from that language and its lexical similarity with Hindi. Each cell in the upper right triangle contains the BLEU scores using a. Our proposed approach (our), b. Multi source ensembling approach (ens), c. Multi Encoder Multi Source approach (me) and d. The lexical similarity (sim; in tiny font size). The best BLEU score is in bold. The train, dev, test split sizes are 45600, 1000 and 2400 lines respectively.}
\end{table*}

\begin{table*}[t]
\small
\begin{center}
\setlength{\tabcolsep}{.30em}
\begin{tabular}{|c|c|c|c|c|c|c|c|c|c|c|}
\hline
\multirow{2}{*}{\textbf{\begin{tabular}[c]{@{}c@{}}Corpus Type\\ \textit{\tiny Train Size}\end{tabular}}}            & \multirow{2}{*}{\textbf{\begin{tabular}[c]{@{}c@{}}Language\\ Pair\end{tabular}}} & \multirow{2}{*}{\textbf{\begin{tabular}[c]{@{}c@{}}BLEU\\ tst2010\end{tabular}}} & \multirow{2}{*}{\textbf{\begin{tabular}[c]{@{}c@{}}BLEU\\ tst2013\end{tabular}}} & \multirow{2}{*}{\textbf{\begin{tabular}[c]{@{}c@{}}Number\\ of sources\end{tabular}}} & \multicolumn{3}{c|}{\textbf{\begin{tabular}[c]{@{}c@{}}BLEU\\ tst2010\end{tabular}}}       & \multicolumn{3}{c|}{\textbf{\begin{tabular}[c]{@{}c@{}}BLEU\\ tst2013\end{tabular}}}      \\ \cline{6-11} 
                                                                                           &                                                                                   &                                                                                  &                                                                                  &                                                                                       & our                            & ens                             & me                     & our                            & ens                            & me                     \\ \hline
\multirow{2}{*}{\textbf{\begin{tabular}[c]{@{}c@{}}3 lingual\\ \textit{\tiny 191381 lines}\end{tabular}}} & \textbf{Fr-En}                                                                    & 19.72                                                                            & 22.05                                                                            & \multirow{2}{*}{\textbf{2}}                                                           & \multirow{2}{*}{\textbf{22.56}} & \multirow{2}{*}{18.64}          & \multirow{2}{*}{22.03} & \multirow{2}{*}{\textbf{24.02}} & \multirow{2}{*}{18.45}         & \multirow{2}{*}{23.92} \\ \cline{2-4}
                                                                                           & \textbf{De-En}                                                                    & 16.19                                                                            & 16.13                                                                            &                                                                                       &                                 &                                 &                        &                                 &                                &                        \\ \hline
\multirow{3}{*}{\textbf{\begin{tabular}[c]{@{}c@{}}4 lingual\\ \textit{\tiny 84301 lines}\end{tabular}}}  & \textbf{Fr-En}                                                                    & 9.02                                                                             & 7.78                                                                             & \multirow{3}{*}{\textbf{3}}                                                           & \multirow{3}{*}{11.70}          & \multirow{3}{*}{\textbf{12.86}} & \multirow{3}{*}{10.30} & \multirow{3}{*}{9.16}           & \multirow{3}{*}{\textbf{9.48}} & \multirow{3}{*}{7.30}  \\ \cline{2-4}
                                                                                           & \textbf{De-En}                                                                    & 7.58                                                                             & 5.45                                                                             &                                                                                       &                                 &                                 &                        &                                 &                                &                        \\ \cline{2-4}
                                                                                           & \textbf{Ar-En}                                                                    & 6.53                                                                             & 5.25                                                                             &                                                                                       &                                 &                                 &                        &                                 &                                &                        \\ \hline
\multirow{4}{*}{\textbf{\begin{tabular}[c]{@{}c@{}}5 lingual\\ \textit{\tiny 45684 lines}\end{tabular}}}  & \textbf{Fr-En}                                                                    & 6.69                                                                             & 6.36                                                                             & \multirow{4}{*}{\textbf{4}}                                                           & \multirow{4}{*}{8.34}           & \multirow{4}{*}{\textbf{9.23}}  & \multirow{4}{*}{7.79}  & \multirow{4}{*}{\textbf{6.67}}  & \multirow{4}{*}{6.49}          & \multirow{4}{*}{5.92}  \\ \cline{2-4}
                                                                                           & \textbf{De-En}                                                                    & 5.76                                                                             & 3.86                                                                             &                                                                                       &                                 &                                 &                        &                                 &                                &                        \\ \cline{2-4}
                                                                                           & \textbf{Ar-En}                                                                    & 4.53                                                                             & 2.92                                                                             &                                                                                       &                                 &                                 &                        &                                 &                                &                        \\ \cline{2-4}
                                                                                           & \textbf{Cs-En}                                                                    & 4.56                                                                             & 3.40                                                                             &                                                                                       &                                 &                                 &                        &                                 &                                &                        \\ \hline
\end{tabular}
\end{center}
\caption{\textbf{IWSLT corpus results}: BLEU scores \label{iwsltresults} for the single source and N source settings using the IWSLT corpus. The languages are French (Fr), German (De), Arabic (Ar), Czech (Cs) and English (En). We give the BLEU scores for two test sets tst2010 and tst2013 which we translate using a. Our proposed approach (our), b. Multi source ensembling approach (ens) and c. Multi Encoder Multi Source approach (me). The best BLEU score is in bold. The train corpus sizes are given in tiny font size. Refer to Table~\ref{iwsltstats} for details on corpora sizes.}
\end{table*}

\begin{table}[!h]
\small
\begin{center}
\setlength{\tabcolsep}{.30em}
\begin{tabular}{|c|c|l|c|c|c|c|}
\cline{1-2} \cline{4-7}
\multirow{2}{*}{\textbf{\begin{tabular}[c]{@{}c@{}}Language\\ Pair\end{tabular}}} & \multirow{2}{*}{\textbf{BLEU}} &  & \multirow{2}{*}{\textbf{\begin{tabular}[c]{@{}c@{}}Source\\ Combination\end{tabular}}} & \multicolumn{3}{c|}{\textbf{BLEU}}       \\ \cline{5-7} 
                                                                                  &                                &  &                                                                                        & our             & ens   & me             \\ \cline{1-2} \cline{4-7} 
\textbf{Es-En}                                                                    & 49.20                          &  & \textbf{Es+Fr}                                                                         & \textbf{49.93*} & 46.65 & 47.39          \\ \cline{1-2} \cline{4-7} 
\textbf{Fr-En}                                                                    & 40.52                          &  & \textbf{Fr+Ru}                                                                         & \textbf{43.99}  & 40.63 & 42.12          \\ \cline{1-2} \cline{4-7} 
\textbf{Ar-En}                                                                    & 40.58                          &  & \textbf{Fr+Ar}                                                                         & 43.85           & 41.13 & \textbf{44.06} \\ \cline{1-2} \cline{4-7} 
\textbf{Ru-En}                                                                    & 38.94                          &  & \textbf{Ar+Ru}                                                                         & 41.66           & 43.12 & \textbf{43.69} \\ \cline{1-2} \cline{4-7} 
\end{tabular}
\end{center}
\caption{\textbf{UN corpus results}: BLEU scores \label{uncorpus2Mresults} for the single source and 2 source settings using the UN corpus. The languages are Spanish (Es), French (Fr), Russian (Ru), Arabic (Ar) and English (En). We give the BLEU scores for for the test set which we translate using a. Our proposed approach (our), b. Multi source ensembling approach (ens)  and c. Multi Encoder Multi Source approach (me). Note that we do not try all language pairs. The highest score is the one in bold. All BLEU score improvements are statistically significant (p \textless 0.001) compared to those obtained using either of the source languages independently. The train, dev, test split sizes are 2M, 4k and 4k lines respectively.}
\end{table}

For the ILCI corpus setting, Table~\ref{ms1t} contains the BLEU scores for all the settings and lexical similarity scores for all combinations of source languages, two at a time. The caption contains a complete description of the table. The last row of Table~\ref{ms1t} contains the BLEU score for all the multi source settings which uses all 5 source languages.\\
For the results of the IWSLT corpus setting, refer to Table~\ref{iwsltresults}. Finally, refer to Table~\ref{uncorpus2Mresults} for the UN corpus setting.

\subsection{Analysis}
\subsubsection{Main findings}
From Tables~\ref{ms1t}, \ref{iwsltresults} and Table~\ref{uncorpus2Mresults} it is clear that our simple source sentence concatenation based approach (under columns labeled ``our") is able to leverage multiple languages leading to significant improvements compared to the BLEU scores obtained using any of the individual source languages. The ensembling (under columns labeled ``ens") and the multi encoder (under columns labeled ``me") approaches also lead to improvements in BLEU. Note that in every single case, gains in BLEU are statistically significant regardless of the methods used. It should be noted that in a resource poor scenario ensembling generally outperforms all other approaches but in a resource rich scenario our method as well as the multi encoder method are much better. However, the comparison with the ensembling method is unfair to our method since the former uses N times more parameters than the latter. However, one important aspect of our approach is that the model size for the multi source systems is the same as that of the single source systems since the vocabulary sizes are exactly the same. The multi encoder systems involve more parameters whereas the ensembling approach does not allow for the source languages to truly interact with each other.
\subsubsection{Correlation between linguistic similarity and gains using multiple sources}
In the case of the ILCI corpus setting, Table~\ref{ms1t}, it is clear that no matter which source languages are combined, the BLEU scores are higher than those given by the single source systems. Marathi and Bengali are the closest to Hindi (linguistically speaking) compared to the other languages and thus when used together they help obtain an improvement of 4.39 BLEU points compared to when Marathi is used as the only source language (24.63).  However it can be seen that combining any of Marathi, Bengali and Telugu with either English or Tamil lead to smaller gains. There is a strong correlation between the gains in BLEU and the lexical similarity. Bengali and English which have the least lexical similarity (0.18) give only a 1.56 BLEU improvement whereas Bengali and Marathi which have the highest lexical similarity (0.46) give a BLEU improvement of 4.42 using our multi-source method. This seems to indicate that although multiple source languages do help, source languages that are linguistically closer to each other are responsible for maximum gains (as evidenced by the correlation between lexical similarity and gains in BLEU). Finally, the last row of Table~\ref{ms1t} shows that using additional languages lead to further gains leading to a BLEU score of 31.56 which is 6.96 points above when only Marathi is used as the only source language and 2.54 points above when Marathi and Bengali are used as the source languages. As future work it will be worthwhile to investigate the diminishing returns in BLEU improvement obtained per additional language.

\subsubsection{Performance in resource rich settings}
In the UN corpus setting, Table~\ref{uncorpus2Mresults}, where we used approximately 2 million training sentences, we also obtained improvements in BLEU. In the case of the single source systems we observed that the BLEU score for Spanish-English was around 9 BLEU points higher than for French-English which is consistent with the observations in the original work concerning the construction of the UN corpus \cite{ZIEMSKI16.1195}. Furthermore, combining using French and Spanish together leads to a small (0.7) improvement in BLEU (over Spanish-English) that is statistically significant (p \textless 0.001) which is to be expected since the BLEU for Spanish-English is already much better than the BLEU for French-English. Since the BLEU scores for French, Arabic and Russian to English are closer to each other we can see that the BLEU scores for French+Arabic, French+Russian and Arabic+Russian to English are around 3 BLEU points higher than those of their respective single source counterparts.However, they do not beat the performance\footnote{The difference in performance between multi-encoder approach and our approache for French+Arabic is not significant.} of the multi-encoder models which have roughly twice the number of parameters.\\
Similar gains in BLEU are observed in the IWSLT corpus setting. Halving the size of the training corpus (from trilingual to 4-lingual) leads to baseline BLEU scores being reduced by half (19.72 to 9.62 for French-English tst2010 test set) but using an additional source leads to a gain of roughly 2 BLEU points. Although the gains are not as high as seen in the ILCI corpus setting it must be noted that the test set for the ILCI corpus is easier in the sense that it contains many short sentences compared to the IWSLT test sets. Our method does not show any gains in BLEU for the tst2013 test set in the 4-lingual setting, an anomaly which we plan to investigate in the future.

\subsection{Studying multi-source attention}
\begin{figure}[!h]
    \centering
  \begin{minipage}[t]{0.35\linewidth}
  \centerline{\includegraphics[width=\linewidth]{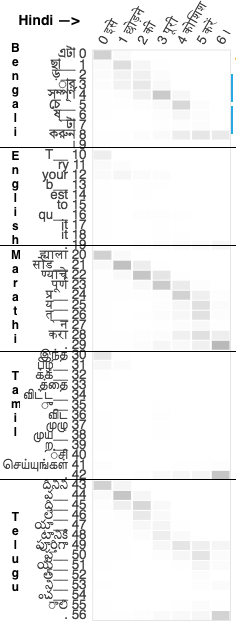}}
    \caption{Attention Visualization for ILCI corpus setting for Bengali, English, Marathi, Tamil and Telugu to Hindi. A horizontal black line is used to separate the source languages but the NMT system receives a single, long multi-source sentence.}
    \label{attvis1}
  \end{minipage}%
  \hfill%
  \begin{minipage}[t]{0.40\linewidth}
  \centerline{\includegraphics[width=\linewidth]{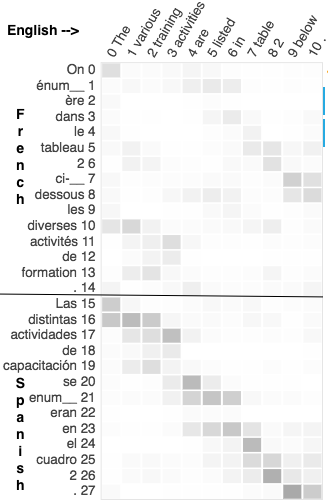}}
    \caption{Attention Visualization for UN corpus setting for French and Spanish to English. A horizontal black line is used to separate the source languages but the NMT system receives a single, long multi-source sentence.}
    \label{attvis3}
    \end{minipage}%
\end{figure}
In order to understand whether or not our multi-source NMT approach prefers certain language over others, we obtained visualizations for the attention vectors for a few sentences from the test set. Refer to Figure~\ref{attvis1} for an example. Firstly, it can be seen that the NMT model learns sentence boundaries although we did not specify delimiters between sentences, Note that, in the figure, we use a horizontal line to separate the languages but the NMT system receives a single, long multi-source sentence. The words of the target sentence in Hindi are arranged from left to right along the columns whereas the words of the multi-source sentence are arranged from top to bottom across the rows. Note that the source languages (and lexical similarity scores with Hindi) are in the following order: Bengali (0.52), English (0.20), Marathi (0.51), Tamil (0.30), Telugu (0.42).\\ The most interesting thing that can be seen is that the attention mechanism  focuses on each language but with varying degrees of focus. Bengali, Marathi and Telugu are the three languages that receive most of the attention (highest lexical similarity scores with Hindi) whereas English and Tamil (lowest lexical similarity scores with Hindi) barely receive any. Building on this observation we believe that the gains we obtained by using all 5 source languages were mostly due to Bengali, Telugu and Marathi whereas the NMT system learns to practically ignore Tamil and English. However there does not seem to be any detrimental effect of using English and Tamil.\\
From Figure~\ref{attvis3} it can be seen that this observation also holds in the UN corpus setting for French+Spanish to English where the attention mechanism gives a higher weight to Spanish words compared to French words since the Spanish-English translation quality is about 9 BLEU points higher than the French-English translation quality. It is also interesting to note that the attention can potentially be used to extract a multilingual dictionary simply by learning a N-source NMT system and then generating a dictionary by extracting the words from the source sentence that receive the highest attention for each target word generated.

\section{Conclusion and Future Work}
In this paper, we have explored a simple approach for ``Multi-Source Neural Machine Translation" that can used with any NMT system seen as a black-box. We have evaluated it in a resource poor as well as a resource rich setting using the ILCI, IWSLT and UN corpora. We have compared our approach with two other previously proposed approaches and showed that it gives competitive results with other state of the art methods while using less than half the number of parameters (for 2 source models). It is domain and language independent and the gains are significant. We also observed, by visualizing attention, that NMT is able to identify sentence boundaries without sentence delimiters and focuses on some languages by practically ignoring others indicating that language relatedness is one of the aspects that should be considered in a multilingual MT scenario. Although we have not explored other multi-source NLP tasks in this paper, we believe that our method and findings will be applicable to them.\\
In the future we plan on exploring the language relatedness phenomenon by considering even more languages. We also plan on investigating the extraction of multilingual dictionaries by analyzing the attention links and on how we can obtain a single NMT model that can translate up to N source languages and thereby function in a situation where some source sentences in certain languages are missing.
\small

\bibliographystyle{apalike}
\bibliography{mtsummit2017}

\end{document}